\def\year{2021}\relax
\documentclass[letterpaper]{article} 
\usepackage{aaai21}  
\usepackage{times}  
\usepackage{helvet} 
\usepackage{courier}  
\usepackage[hyphens]{url}  
\usepackage{graphicx} 
\usepackage{graphicx}  
\usepackage{natbib}  
\usepackage{caption} 
\title{ACSNet: Action-Context Separation Network for Weakly Supervised\\Temporal Action Localization}
\author{
   Ziyi Liu\textsuperscript{\rm 1}, 
   Le Wang\textsuperscript{\rm 1}\thanks{Corresponding author.}, 
   Qilin Zhang\textsuperscript{\rm 2},
   Wei Tang\textsuperscript{\rm 3},
   Junsong Yuan\textsuperscript{\rm 4},
   Nanning Zheng\textsuperscript{\rm 1},
   Gang Hua\textsuperscript{\rm 5}
}
\affiliations{

   \textsuperscript{\rm 1}Institute of Artificial Intelligence and Robotics, Xi`an Jiaotong University~
   \textsuperscript{\rm 2}HERE Technologies \\
   \textsuperscript{\rm 3}University of Illinois at Chicago~
   \textsuperscript{\rm 4}The State University of New York at Buffalo~
   \textsuperscript{\rm 5}Wormpex AI Research \\
   liuziyi@stu.xjtu.edu.cn, 
   \{lewang, nnzheng\}@mail.xjtu.edu.cn,\\
   tangw@uic.edu,
   jsyuan@buffalo.edu,
   \{samqzhang, ganghua\}@gmail.com,

}

\newcommand{\name}{ACSNet}
\newcommand{\fb}{FB branch}
\newcommand{\ac}{AC branch}

\newcommand{\lzynet}{CleanNet}

\newcommand{\wtal}{WS-TAL}

\def\ie{\emph{i.e}.}
\def\etal{\emph{et al}.}
\def\eg{\emph{e.g}.}
\usepackage{times}
\usepackage{epsfig}
\usepackage{graphicx}
\usepackage{amsmath}
\usepackage{amssymb}
\usepackage{subcaption}
\usepackage{placeins}
\usepackage{multirow}
\usepackage{array}
\usepackage{diagbox}
\usepackage{multirow}
\usepackage[switch]{lineno}  %
\begin{document}
\maketitle

\begin{abstract}
The object of Weakly-supervised Temporal Action Localization (WS-TAL) is to localize all action instances in an untrimmed video with only video-level supervision. Due to the lack of frame-level annotations during training, current WS-TAL methods rely on attention mechanisms to localize the foreground snippets or frames that contribute to the video-level classification task. This strategy frequently confuse context with the actual action, in the localization result. Separating action and context is a core problem for precise WS-TAL, but it is very challenging and has been largely ignored in the literature. In this paper, we introduce an Action-Context Separation Network (\name) that explicitly takes into account context for accurate action localization. It consists of two branches (\ie, the Foreground-Background branch and the Action-Context branch). The Foreground-Background branch first distinguishes foreground from background within the entire video while the Action-Context branch further separates the foreground as action and context. We associate video snippets with two latent components (\ie, a positive component and a negative component), and their different combinations can effectively characterize foreground, action and context. Furthermore, we introduce extended labels with auxiliary context categories to facilitate the learning of action-context separation. Experiments on THUMOS14 and ActivityNet v1.2/v1.3 datasets demonstrate the \name~outperforms existing state-of-the-art WS-TAL methods by a large margin.
\end{abstract}

\section{Introduction}
\begin{figure}[t]
  \begin{center}
  \includegraphics[width=0.48\textwidth]{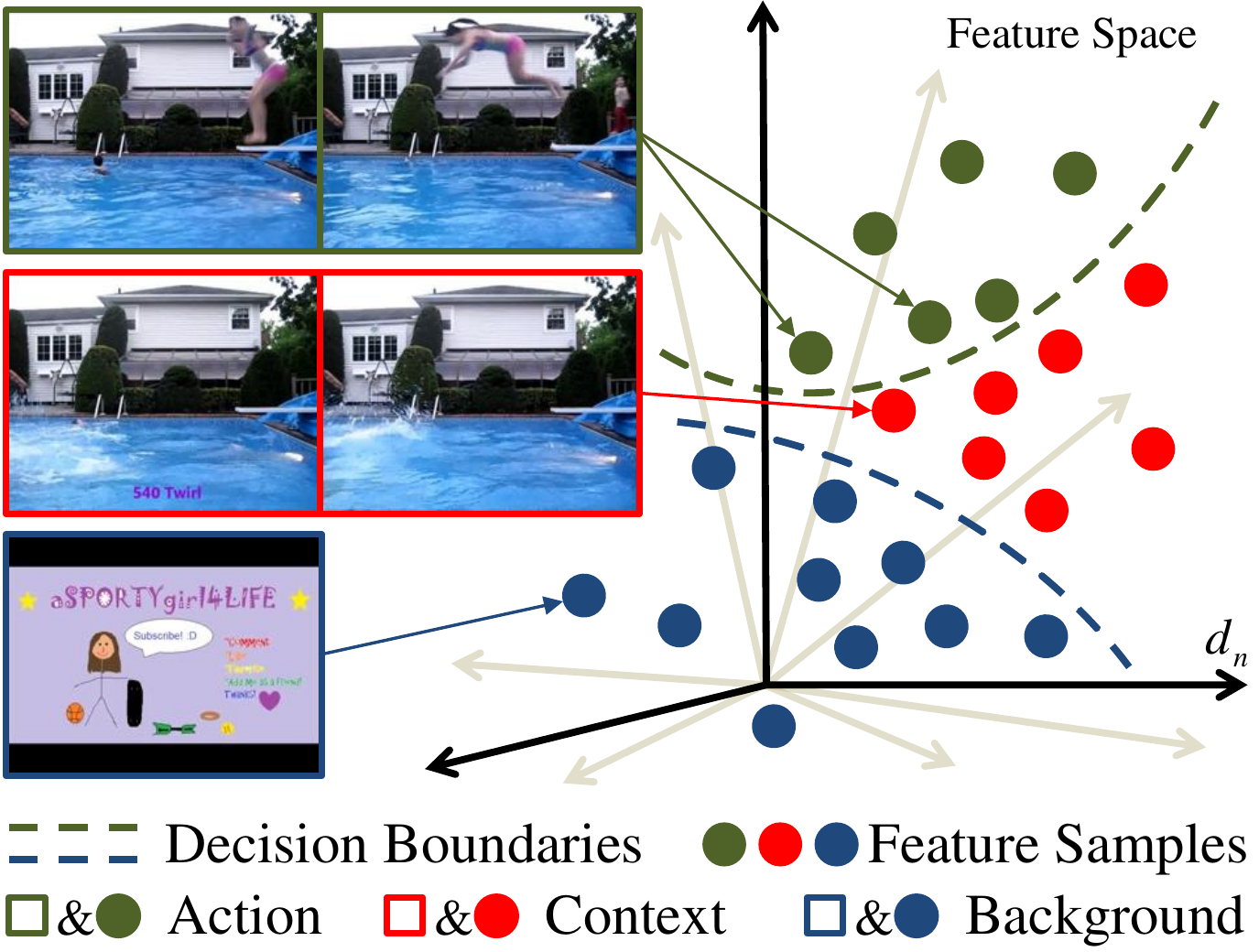}\\
  \caption{The illustration of action, context and background in terms of frames and points in feature space. The green dashed line is the desired boundary for the localization task. However, based on the given video-level categorical labels, the blue dashed line is learned, due to the high co-occurrence and visual similarity of action and context.
Existing methods frequently identify both red and green dots as actions. The main challenge in WS-TAL is how to isolate context from action instances with merely video-level categorical labels
  }\label{fig:fig1}
  \end{center}
\end{figure}

\begin{figure}[t]
  \begin{center}
  \includegraphics[width=0.48\textwidth]{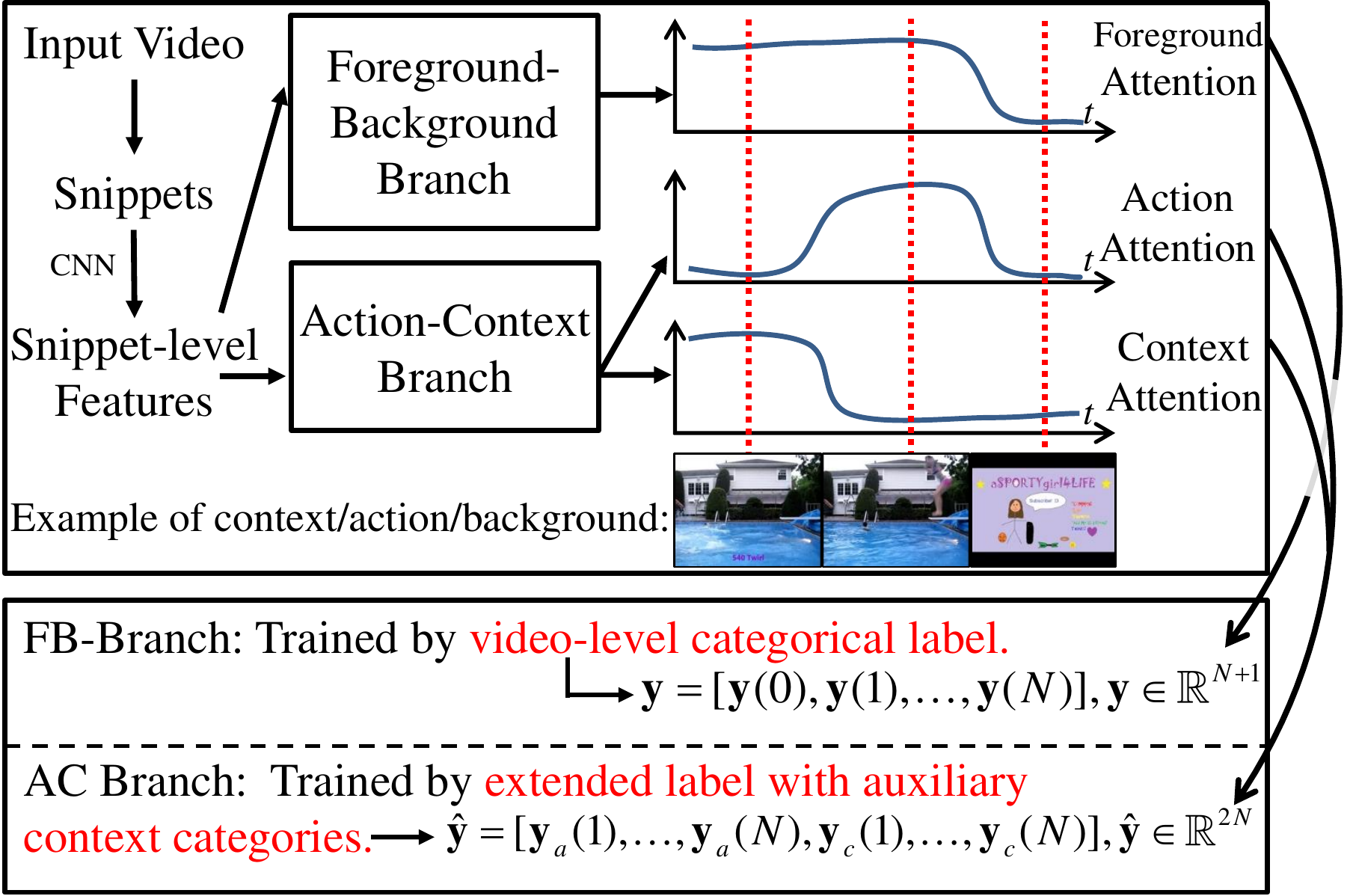}\\
  \caption{An overview of our main idea, \ie, using extended label with auxiliary context categories to guide the training of action/context attentions. Unfortunately, such an idea is nontrivial to implement due to ``lack of explicit action-context constraint'' and ``lack of explicit supervision''.
  }\label{fig:framework_simp}
  \end{center}
\end{figure}

\noindent Temporal Action Localization (TAL) aims to localize temporal starts and ends of specific action categories in a video. It serves as a fundamental tool for several practical applications such as action retrieval, intelligent surveillance and video summarization~\cite{lee2012discovering,vishwakarma2013survey,surveyDeepAction,Kang2016Review,yao2019review}.
Although fully supervised TAL methods have recently achieved remarkable progress~\cite{buch2017sst,xu2017RC3D,gao2017turn,xu2017RC3D,chao2018rethinking,lin2018bsn,lin2019bmn,zeng2019graph}, manually annotating the precise temporal boundaries of action instances in untrimmed videos is time-consuming and challenging. This limitation motivates the weakly supervised setting where only video-level categorical labels are provided for model training. Compared with temporal boundary annotations, video-level categorical labels are easier to collect, and they help avoid the localization bias introduced by human annotators.

Existing weakly-supervised temporal action localization (WS-TAL) methods~\cite{wang2017untrimmednets,nguyen2017weakly,WTALC,nguyen2019weakly} leverage attention mechanisms to categorize snippets or sampled frames into foreground and background based on their contribution to the video-level classification task, \ie, to find the blue dashed line in Figure~\ref{fig:fig1}.
Then temporal action localization is reformulated as selecting consecutive foreground snippets belonging to each category.
However, the foreground localized through video-level categorization involves not only the actual action instance but also its surrounding \emph{context}. As illustrated in Figure~\ref{fig:fig1}, context is snippets or frames that frequently \textit{co-occur} with the action instances of a specific category but should not be included in their localization. Different from background, which is class-agnostic, context provides strong evidence for action classification and thus can be easily confused with the action instances.
We believe separating the action instances and their context is a core problem in WS-TAL, and it is very challenging due to the co-occurrence nature.

The goal of this paper is to address the action-context separation (ACS) problem in the weakly-supervised setting so as to achieve more precise action localization. We first introduce auxiliary context categories for each action class during training. As shown in Figure~\ref{fig:framework_simp}, each video-level category is divided into two sub-categories, respectively corresponding to the actual action and its context.
Prior methods exploit foreground attention to achieve foreground-background separation. However, this simple idea is not applicable to action-context separation due to two difficult issues. (1) Lack of explicit action-context constraints: The sum-to-one constraint ~\cite{nguyen2019weakly} of the foreground and background attention scores does not apply to action-context separation. (2) Lack of explicit supervision: Both action and context can contribute to action classification, so the only available video-level categorical labels cannot provide direct supervision for them.

\begin{figure*}[t]
\setlength{\belowcaptionskip}{-0.2cm} 
  \begin{center}
  \includegraphics[width=1\textwidth]{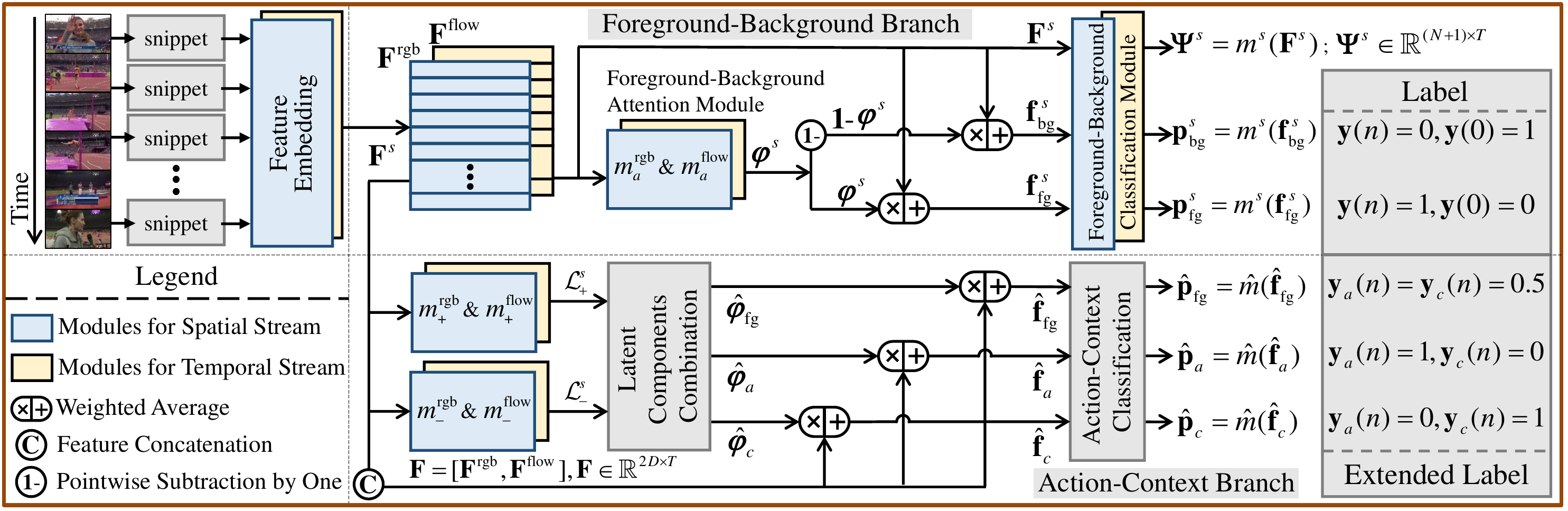}\\
  \caption{The framework of the proposed ACSNet, which has two branches, \ie, Foreground-Background branch and Action-Context branch. The input video is first processed by the feature embedding to get features from both spatial and temporal streams. The \fb~focuses on foreground-background separation while the \ac~focuses on action-context separation. Video-level labels are extended to facilitate the action-context separation.
  }\label{fig:framework}
  \end{center}
\end{figure*}

To address these two difficult issues, we introduce the Action-Context Separation Network (\name). As illustrated in Figure~\ref{fig:framework}, it consists of two branches, \ie, the Foreground-Background branch (\fb) and the Action-Context branch (\ac).
The \fb~divides an untrimmed video into foreground and background based on whether a snippet supports the video-level classification. This is achieved via snippet-level categorical predictions (SCPs) and snippet-level attention predictions (SAPs), \eg, foreground attention in Figure~\ref{fig:framework_simp}.
Subsequently, the \ac~further divides the obtained foreground into action and context by associating each video snippet with two latent components, \ie, a positive component and a negative component.
Different combinations of these two components respectively characterize the foreground, action and context. This enables effective action-context separation with only video-level supervision.
Finally, the output of \ac~facilitates the TAL by providing (1) temporal action proposals with more accurate boundaries and (2) more reliable proposal confidence scores.

The contribution of this paper is summarized below.
\begin{enumerate}
    \item Prior WS-TAL approaches take it for granted that the foreground localized via the classification attention is equivalent to the actual action instance, and thus they unavoidably include the co-occurring context in the localization result. We address this challenge via a novel action-context separation network (ACSNet), which not only distinguishes foreground from background but also separates action and context within the foreground to achieve more precise action localization.
\item The proposed ACSNet features a novel Action-Context branch. It can individually characterize foreground, action and context using different combinations of two latent components, \ie, the positive component and the negative component.
\item We propose novel extended labels with auxiliary context categories. By explicitly decoupling the actual action and its context, this new representation facilitates effective learning of action-context separation.
\item Extensive experimental results indicate the proposed ACSNet can effectively perform action-context separation. It significantly outperforms state-of-the-art methods
    on three benchmarks, and it is even comparable to recent fully-supervised methods.
\end{enumerate}

\section{Related Work of WS-TAL}\label{sec:TALweakSuper}
Different from action recognition which is essentially a classification task~\cite{feichtenhofer2016convolutional,Twostream,TSN,ji20133d,sun2015human,tran2015learning,slowfast}, TAL requires finer-grained predictions with temporal boundaries of the target action instances. WS-TAL methods address it without temporal annotations, which is first introduced in \cite{sun2015temporal}. To distinguish action instances from background, the attention mechanism is widely adopted for foreground-background separation.
UntrimmedNet~\cite{wang2017untrimmednets} formulates the attention mechanism as a soft selection module to localize target action, and the final localization is achieved by thresholding the snippets' action scores.
STPN~\cite{nguyen2017weakly} proposes a sparsity loss based on the soft selection module of UntrimmedNet, which can facilitate the selection of action instances.
Nguyen \etal~\cite{nguyen2019weakly} characterize background by an additional background loss and introduce other losses to guide the attention.
For better evaluation of temporal action proposals,
W-TALC~\cite{WTALC} proposes a co-activity loss to enforce the feature similarity among localized instances. AutoLoc~\cite{shou2018autoloc} uses an ``outer-inner-contrastive loss'' to predict and regress temporal boundaries.
Liu \etal~\cite{Liu_2019_CVPR} exploit a multi-branch neural network to discover distinctive action parts and fuse them to ensure completeness.
CleanNet~\cite{lzyiccv} designs a ``contrast score'' by leveraging temporal contrast in SCPs to achieve end-to-end
training of localization.

However, driven by the video-level classification labels, the existing attention mechanism is merely able to capture the difference between foreground and background for classification, instead of action and non-action for localization. The proposed \name~manages to distinguish action instances from their surrounding context, and we extend labels by introducing auxiliary context categories to make the framework trainable.

\section{Action-Context Separation Network}\label{sec:method}

In this section, we introduce the extended video-level labels with auxiliary context categories (Section~\ref{sec:label}) and the proposed Action-Context Separation Network (\name). As illustrated in Figure~\ref{fig:framework}, the \name~consists of two branches, \ie, Foreground-Background branch (\fb) and Action-Context branch (\ac). After feature extraction from the given video (Section~\ref{sec:feature}), \fb~distinguishes the foreground from background (Section~\ref{sec:fb}). The obtained foreground contains both action and context. Subsequently, \ac~localizes the actual temporal action instances by performing action-context separation within the foreground (Section~\ref{sec:ac}). To guide the training of ACS, additional losses are introduced (Section~\ref{sec:loss}).

\subsection{Extending Video-Level Labels}\label{sec:label}

Suppose we are given a video $V$ with a video-level categorical label $\mathbf{y}=[\mathbf{y}(0), \mathbf{y}(1), \dots, \mathbf{y}(N)]$, where $\mathbf{y}(n)=1$ if $V$ contains the $n$-th action category. $N$ is the total number of action categories, $\mathbf{y}(0)$ represents the background category.
To guide the division of foreground into action and context, we extend $\mathbf{y}$ with auxiliary context categories as
\begin{equation}\label{eq:label}\small
\mathbf{\widehat{y}}=[\mathbf{y}_a(1), \dots,  \mathbf{y}_a(N),~\mathbf{y}_c(1), \dots,  \mathbf{y}_c(N)],~\mathbf{\widehat{y}} \in \mathbb{R}^{2N},
\end{equation}
where $\mathbf{y}_a(n)$ and $\mathbf{y}_c(n)$ denote the $n$-th action category and its corresponding context, respectively.
As shown in Figure~\ref{fig:framework}, $\mathbf{y}\in \mathbb{R}^{N+1}$ is used in \fb~and $\mathbf{\widehat{y}}\in \mathbb{R}^{2N}$ is used in \ac.

\subsection{Baseline Modules}

This section introduces the baseline modules used in \name, including feature extraction and \fb~based on the attention mechanism. While they are not our main contribution, we introduce them for completeness. Similar modules have been explored and adopted by existing methods~\cite{nguyen2017weakly, WTALC, nguyen2019weakly, lee2019background}.

\subsubsection{Feature Extraction}\label{sec:feature}
The input of the feature extraction module is the given video $V=\{s_t\}_{t=1}^T$, which is divided into $T$ non-overlapping snippets. The outputs are the corresponding features of each snippet.
For each snippet $s_t$, the corresponding $D$-dimensional features are extracted from two streams, \ie, the spatial stream (RGB) and the temporal stream (optical flow), denoted as $\mathbf{F}^\textrm{rgb}(t) \in \mathbb{R}^{D}$ and $\mathbf{F}^\textrm{flow}(t) \in \mathbb{R}^{D}$, respectively. Afterwards, the video $V$ is represented as $\mathbf{F}^\textrm{rgb} \in \mathbb{R}^{D \times T}$ and $\mathbf{F}^\textrm{flow} \in \mathbb{R}^{D \times T}$.

For notational simplicity, we use superscript ``$s$'' to indicate the notations used in both streams in the rest of the paper. The notations of the spatial/temporal stream can be obtained by substituting the superscript ``$s$'' with ``\textrm{rgb}/\textrm{flow}''.
For example, $\mathbf{F}^{s}$ can represent either $\mathbf{F}^\textrm{rgb}$ or $\mathbf{F}^\textrm{flow}$.
\subsubsection{Foreground-Background Branch}\label{sec:fb}

The goal of the \fb~is to divide the entire video into two parts, \ie, foreground and background, which can be trained by the video-level categorical label $\mathbf{y}=[\mathbf{y}(0), \mathbf{y}(1), \dots,  \mathbf{y}(N)]$.

The inputs of \fb~are the features $\mathbf{F}^{s} \!\in\! \mathbb{R}^{D \times T}$, and the outputs are the snippet-level attention predictions (SAPs, $
{\pmb{\varphi}}\!\in\! \mathbb{R}^{1 \times T}$) and the snippet-level classification predictions (SCPs, $
{\pmb{\Psi}} \!\in\! \mathbb{R}^{(N\!+\!1) \times T}$).
Accordingly, \fb~consists of two sub-modules, \ie, attention module ($m_{a}^s$) and Foreground-Background classification module ($m^s$).
The SAPs and SCPs of each stream are obtained by
\begin{linenomath}\begin{align}\label{eq:sap_s}
&\pmb{\varphi}^{\textrm{s}}=m_{a}^s(\mathbf{F}^s),~~\pmb{\varphi}^{\textrm{s}}\in \mathbb{R}^{1 \times T},\\
&\pmb{\Psi}^{{s}} = m^s(\mathbf{F}^s), ~~\pmb{\Psi}^{{s}} \in \mathbb{R}^{(N+1) \times T}.
\end{align}\end{linenomath}
Subsequently, the outputs of two streams are weighted to get the final SAPs and SCPs as
\begin{linenomath}\begin{align}\label{eq:fuse}
&{\pmb{\varphi}}=\alpha\pmb{\varphi}^{\textrm{rgb}}+(1-\alpha)\pmb{\varphi}^{\textrm{flow}},\\
&{\pmb{\Psi}} = \alpha\pmb{\Psi}^{\textrm{rgb}} + (1-\alpha)\pmb{\Psi}^{\textrm{flow}},
\end{align}\end{linenomath}
where $\alpha=0.5$ by default in our experiments.
We implement $m_{a}^s$ with a fully-connected (FC) layer followed by a sigmoid activation function. And $m^s$ is implemented by an FC layer.

To train $m_{a}^s$ and $m^s$ with only video-level label, video-level prediction is needed. Therefore, we calculate the video-level foreground feature as
\begin{equation}\label{fg}
\mathbf{f}_\textrm{fg}^s =\frac{1}{T}\sum_{t=1}^{T}\pmb{\varphi}^{{s}}(t)\mathbf{F}^s(t),~\mathbf{f}_\textrm{fg}^s \in \mathbb{R}^D.
\end{equation}
Similarly, the video-level background feature is obtained by
\begin{equation}\label{bg}
\mathbf{f}_\textrm{bg}^s =\frac{1}{T}\sum_{t=1}^{T}(1-\pmb{\varphi}^{{s}}(t))\mathbf{F}^s(t),~\mathbf{f}_{bg}^s \in \mathbb{R}^D.
\end{equation}
After obtaining $\mathbf{f}_\textrm{fg}^s$ and $\mathbf{f}_\textrm{bg}^s$, we feed them into $m^s$ to obtain the video-level prediction, \ie,
the foreground prediction ($\mathbf{p}_\textrm{fg}^s \in \mathbb{R}^{N+1}$) and background prediction ($\mathbf{p}_\textrm{bg}^s \in \mathbb{R}^{N+1}$), defined as
\begin{linenomath}\begin{align}\label{eq:p_fgbg}
\mathbf{p}_\textrm{fg}^s=m^s(\mathbf{f}_\textrm{fg}^s),~
\mathbf{p}_\textrm{bg}^s=m^s(\mathbf{f}_\textrm{bg}^s).
\end{align}\end{linenomath}

Given video-level predictions in Eq.(\ref{eq:p_fgbg}), the \fb~can be trained via regular cross-entropy loss.
For $\mathbf{p}_\textrm{fg}^s$, its label is $\mathbf{{y}}$, where $\mathbf{y}(n)=1$ if $V$ contains the $n$-th action category, as shown in Figure~\ref{fig:framework}.
While for $\mathbf{p}_\textrm{bg}^s$, assuming that all videos contain background snippets, its label is always $\mathbf{y}(0)=1$ and $\mathbf{y}(n)=0,~n=1,2,...N$.
\subsection{Action-Context Branch}\label{sec:ac}

The attention mechanism trained by $\mathbf{y}$ will be distracted by context because both action and context can support video-level classification.
To avoid such distraction, after distinguishing the foreground from background, we further separate action and context within the foreground to locate the actual action instances in this section.

The inputs of the \ac~are features from two streams ($\mathbf{F}^s$ obtained in Section~\ref{sec:feature}) and SAPs (${\pmb{\varphi}}$ obtained in Section~\ref{sec:fb}).
The \ac~consists of three sub-modules, \ie, latent components generation, latent components combination, and action-context separation.

\subsubsection{Latent Components Generation.}
We introduce the concept of positive component ($\mathcal{L}^ {s}_{+} \in \mathbb{R}^{1 \times T}$) and negative component ($\mathcal{L}^ {s}_{-} \in \mathbb{R}^{1 \times T}$) to characterize foreground, action and context. Assuming the foreground is represented by two latent components, we define the one corresponding to the actual action as positive component, while the other one as negative component.
They are obtained similarly as the SAPs in Eq.(\ref{eq:sap_s}), by feeding features into positive module ($m_+^s$) and negative module ($m_-^s$)
\begin{linenomath}\begin{align}\label{}
&\mathcal{L}^ {s}_{+}  = m_+^s (\mathbf{F}^s),~
\mathcal{L}^ {s}_{-}  = m_-^s (\mathbf{F}^s).
\end{align}\end{linenomath}
$m_+^s$ and $m_-^s$ share the same architecture (parameters are not shared), with two temporal convolution (Conv1d) layers followed by a ReLU and a sigmoid activation function for the first and the second layer, respectively.

\subsubsection{Latent Components Combination.}
Given $\mathcal{L}^ {s}_{+}$ and $\mathcal{L}^ {s}_{-}$, we use the combination of them to construct the snippet-level foreground attention ($\widehat{\pmb{\varphi}}_{\textrm{fg}} \in \mathbb{R}^{1 \times T}$), action attention ($\widehat{\pmb{\varphi}}_{\textrm{a}} \in \mathbb{R}^{1 \times T}$), and context attention ($\widehat{\pmb{\varphi}}_{\textrm{c}} \in \mathbb{R}^{1 \times T}$).
Specifically, for each stream, we have
\begin{linenomath}\begin{align}
&\widehat{\pmb{\varphi}}_{\textrm{fg}}^s=\sigma(\mathcal{L}^ {s}_{+} + \mathcal{L}^ {s}_{-}),\\
&\widehat{\pmb{\varphi}}_{\textrm{a}}^s=\sigma(\mathcal{L}^ {s}_{+}),\\
&\widehat{\pmb{\varphi}}_{\textrm{c}}^s=\sigma(\mathcal{L}^ {s}_{-} - \mathcal{L}^ {s}_{+}),\label{eq:context}
\end{align}\end{linenomath}
where $\sigma(\cdot)$ denotes the sigmoid function.
Subsequently, the outputs from two streams are fused by weighted average similar to Eq.(\ref{eq:fuse}),
\begin{linenomath}\begin{align}\label{eq:fai}
\widehat{\pmb{\varphi}}_{{z}} = \alpha\widehat{\pmb{\varphi}}_{{z}}^{\textrm{rgb}} + (1-\alpha)\widehat{\pmb{\varphi}}_{{z}}^{\textrm{flow}},~
\widehat{\pmb{\varphi}}_{{z}} ~z \in \{ \textrm{fg}, a, c\},
\end{align}\end{linenomath}
where $\widehat{\pmb{\varphi}}_{{z}} \in \mathbb{R}^{1 \times T}$.
For notational simplicity, we use subscript ``$z$'' to denote either ``$\textrm{fg}$'', ``$a$'' or``$c$'' if necessary. By substituting the subscript ``$z$'' with ``$\textrm{fg}$/$a$/$c$'', $\widehat{\pmb{\varphi}}_{\textrm{fg}}$/$\widehat{\pmb{\varphi}}_{\textrm{a}}$/$\widehat{\pmb{\varphi}}_{\textrm{c}}$ are obtained following Eq.(\ref{eq:fai}).

Instead of directly imposing simple constrains like foreground and background following~\cite{nguyen2019weakly}, \ie, $\widehat{\pmb{\varphi}}_{\textrm{c}}^s=1 - \widehat{\pmb{\varphi}}_{\textrm{a}}^s$, we adopt the combinations of $\mathcal{L}^ {s}_{+}$ and $\mathcal{L}^ {s}_{-}$ to characterize $\widehat{\pmb{\varphi}}_{\textrm{a}}^s$ and $\widehat{\pmb{\varphi}}_{\textrm{c}}^s$ individually. We compared different approaches to obtain $\widehat{\pmb{\varphi}}_{\textrm{c}}^s$ in supplementary material.

\subsubsection{Action-Context Separation.}
After obtaining $\widehat{\pmb{\varphi}}_{\textrm{fg}}$, $\widehat{\pmb{\varphi}}_{\textrm{a}}$ and $\widehat{\pmb{\varphi}}_{\textrm{c}}$, we can start the action-context separation by leveraging label with auxiliary context categories (\ie, $\mathbf{\widehat{y}} \in \mathbb{R}^{2N}$ introduced in Section~\ref{sec:label}). First of all, we select all temporal indices corresponding to foreground snippets as
\begin{equation}\label{}
\mathbb{I}=\{ t~|~{\pmb{\varphi}}(t)>0.5\},~|\mathbb{I}|= T',
\end{equation}
where $|\cdot|$ denotes the cardinality (number of elements).
Subsequently, the video-level feature representations of foreground, action and context are obtained as
\begin{equation}
\widehat{\mathbf{f}}_{z} = \frac{1}{T'}\sum_{t \in \mathbb{I}} \widehat{\pmb{\varphi}}_{{z}}(t)\mathbf{F}(t),~z \in \{ \textrm{fg}, a, c\},\label{eq:f_fg_a_c}
\end{equation}
where $\widehat{\mathbf{f}}_{z} \in \mathbb{R}^{2D \times 1}$ and $\mathbf{F}(t) = \langle\mathbf{F}^\textrm{rgb}(t),\mathbf{F}^\textrm{flow}(t)\rangle~(\mathbf{F}(t) \!\in\! \mathbb{R}^{2D \times 1})$ is the concatenated feature from both streams and $\langle\cdot\rangle$ means concatenation.
By substituting the subscript ``$z$'' with ``$\textrm{fg}$/$a$/$c$'', $\widehat{\mathbf{f}}_\textrm{fg}$, $\widehat{\mathbf{f}}_{a}$ and $\widehat{\mathbf{f}}_{c}$ are calculated following Eq.(\ref{eq:f_fg_a_c}).
Afterwards, they are fed into the action-context classification module $\widehat{m}$ to get the video-level action-context prediction as
\begin{linenomath}\begin{align}\label{}
\widehat{\mathbf{p}}_{z} =\widehat{m}(\widehat{\mathbf{f}}_{z}), \widehat{\mathbf{p}}_{z} \in \mathbb{R}^{2N},~z \in \{ \textrm{fg}, a, c\}.
\end{align}\end{linenomath}
Different from the video-level prediction from \fb~(\ie, $\mathbf{p}_\textrm{fg}^s \in \mathbb{R}^{N+1}$ in Eq.(\ref{eq:p_fgbg})), $\widehat{\mathbf{p}}_{z} \in \mathbb{R}^{2N}$ provides predictions on both action and context categories. Specifically, if the video contains the $n$-th category, the label for $\widehat{\mathbf{p}}_\textrm{fg}$ is $\mathbf{\widehat{y}}=[\mathbf{y}_a(1), \dots, \mathbf{y}_a(N),~\mathbf{y}_c(1), \dots, \mathbf{y}_c(N)]$, where $\mathbf{y}_a(n)=\mathbf{y}_c(n)=0.5$. While for $\widehat{\mathbf{p}}_a$ and $\widehat{\mathbf{p}}_c$, the labels are ($\mathbf{y}_a(n)=1, \mathbf{y}_c(n)=0$) and ($\mathbf{y}_a(n)=0, \mathbf{y}_c(n)=1$), respectively, as shown in Figure~\ref{fig:framework}. After obtaining $\widehat{\mathbf{p}}_z$ and the corresponding labels, the \ac~is also trained via regular cross-entropy loss.

Applying $\widehat{m}$ to each snippet, the snippet-level action-context predictions are obtained as
\begin{equation}\label{}
{\pmb{\Psi}}' = \widehat{m}(\mathbf{F}),~{\pmb{\Psi}}' \in \mathbb{R}^{2N \times T},
\end{equation}
where $\mathbf{F} \in \mathbb{R}^{2D \times T}$ is the concatenated feature. ${\pmb{\Psi}}'$ is leveraged to promote the action and suppress the context, by defining an ``action-context offset ($\widehat{\pmb{\Psi}}\!\in\!\mathbb{R}^{N \times T}$)'' as
\begin{equation}\label{}
\widehat{\pmb{\Psi}}(n,t) \!=\!
\begin{cases}
{\pmb{\Psi}}'(n,t)\!-\!{\pmb{\Psi}}'(2n,t) ~~~~\text{if } t \!\in\! \mathbb{I},\\
0~~~~~~~~~~~~~~~~~~~~~~~otherwise,
\end{cases}
\end{equation}
where ${\pmb{\Psi}}'(n,t)$ (or ${\pmb{\Psi}}'(2n,t)$) is the prediction of the $n$-th action (or corresponding context) of the $t$-th snippet. Intuitively,
$\widehat{\pmb{\Psi}}(n,t)$ means ``offsets'' for the $n$-th class of the $t$-th snippet, compared the prediction of action (${\pmb{\Psi}}'(n,t)$) with context (${\pmb{\Psi}}'(2n,t)$).

In summery, the \ac~outputs snippet-level action score ($\widehat{\pmb{\varphi}}_{\textrm{a}} \in \mathbb{R}^{1 \times T}$) and the action-context offset ($\widehat{\pmb{\Psi}}\in \mathbb{R}^{N \times T}$) for the subsequent localization task.

\subsection{Additional Losses}\label{sec:loss}
In addition to the regular cross-entropy losses, more constrains are required to train the \name~successfully, since there are neither temporal annotations nor action/context annotations available. In this section, we introduce two additional losses to provide extra guidance for \name~training, \ie, $L_g$ and $L_\textrm{mse}$.

For guidance loss $L_g$, due to the lack of ground truth labeled action or context categories, confusion between action and context (\eg, $\widehat{\pmb{\varphi}}_{{a}}$ and $\widehat{\pmb{\varphi}}_{{c}}$, $\mathcal{L}^ {s}_{+}$ and $\mathcal{L}^ {s}_{-}$) will occur due to symmetry. Therefore, additional guidance should be introduced to distinguish action from context, which is achieved by minimizing $L_g$. Specifically, the differences between two streams are leveraged. We adopt weighted binary logistic regression loss function $L_{r}$ to guide $\widehat{\pmb{\varphi}}_{{a}}$ and $\widehat{\pmb{\varphi}}_{{c}}$, where $L_{r}(\mathbf{p},\mathbf{q})$ is denoted as
\begin{equation}\small
L_{r}(\mathbf{p},\mathbf{q})\!=\!-\sum_{i=1}^{l} \left (\frac{q_i\cdot log(p_i)}{l^+} \!+\! \frac{(1\!-\!q_i) \cdot log(1\!-\!p_i)}{l^-} \right ),
\end{equation}
where $\mathbf{p},\mathbf{q} \in \mathbb{R}^{1 \times l}$ and $\mathbf{q}$ is a binary vector indicating positive and negative samples (snippets). $\mathbf{p}$ is the prediction to be regressed. $l^+=\sum q_i$ and $l^-=\sum (1-q_i)$. For action attention $\widehat{\pmb{\varphi}}_{{a}}$, positive time index set ($P_a$) and negative time index set ($N_a$) are defined as
\begin{linenomath}\begin{align}\label{}
&P_a=\{t~|~\widehat{\pmb{\varphi}}_{{a}}^\textrm{rgb}(t)>\theta_h ~\&~ \widehat{\pmb{\varphi}}_{{a}}^\textrm{flow}(t)>\theta_h \},\\
&N_a=\{t~|~\widehat{\pmb{\varphi}}_{{a}}^\textrm{rgb}(t)<\theta_l ~\&~ \widehat{\pmb{\varphi}}_{{a}}^\textrm{flow}(t)<\theta_l \},
\end{align}\end{linenomath}
where $\theta_h$ and $\theta_l$ indicate high and low thresholds, respectively. Intuitively, the snippets with high/low attentions on both streams are regarded as positive/negative samples for action snippets.
For context attention $\widehat{\pmb{\varphi}}_{{c}}$, we assume context contains scenes (excluding action instances), so that the corresponding positive/negative snippet index sets are defined as
\begin{linenomath}\begin{gather}\label{}
P_c=\{t~|~\widehat{\pmb{\varphi}}_{{a}}^\textrm{rgb}(t)>\theta_h ~\&~ \widehat{\pmb{\varphi}}_{{a}}^\textrm{flow}(t)<\theta_l \},\\
N_c = P_a \cup N_a.
\end{gather}\end{linenomath}
Subsequently, the guidance loss $L_g$ is calculated as
\begin{linenomath}\begin{gather}\label{}
\nonumber L_g = L_r(\pmb{\varphi}_{{a}}', [\mathbf{1}(|P_a|), \mathbf{0}(|N_a|)]) + \\
~~~~~~~L_r(\pmb{\varphi}_{{c}}', [\mathbf{1}(|P_c|), \mathbf{0}(|N_c|)]), \\
\pmb{\varphi}_{{a}}' = \langle\widehat{\pmb{\varphi}}_{{a}}(P_a), \widehat{\pmb{\varphi}}_{{a}}(N_a)\rangle,~{\pmb{\varphi}}_{{c}}' = \langle\widehat{\pmb{\varphi}}_{{c}}(P_c), \widehat{\pmb{\varphi}}_{{c}}(N_c)\rangle.
\end{gather}\end{linenomath}
where $\mathbf{1}(d)$ (or $\mathbf{0}(d)$) indicates a d-dimensional vector filled with ones (or zeros).

For $L_\textrm{mse}$, in order to encourage the two latent components to focus on the foreground, we adopt the Mean Squared Error (MSE) loss between $\widehat{\pmb{\varphi}}_\textrm{{fg}}$ and ${\pmb{\varphi}}$, denoted as
\begin{equation}
L_\textrm{mse} = \textrm{MSE}(\widehat{\pmb{\varphi}}_\textrm{{fg}}, G({\pmb{\varphi}})),
\end{equation}
where $G(\cdot)$ is a Gaussian smoothing function.
Finally, the \ac~is trained by minimizing the total loss $L$, calculated as
\begin{equation}\label{}
L=L_\textrm{cls} + \lambda(L_\textrm{mse} + L_g),
\end{equation}
where $L_\textrm{cls}$ is the sum of cross-entropy losses mentioned in Section~\ref{sec:ac}. $\lambda$ is the balancing weight set as $1$.
\section{Localization}

After the inference, \fb~outputs SAPs (${\pmb{\varphi}} \in \mathbb{R}^{1 \times T}$), SCPs (${\pmb{\Psi}} \in \mathbb{R}^{(N+1) \times T}$) and \ac~outputs action score ($\widehat{{\pmb{\varphi}}}_a \in \mathbb{R}^{1 \times T}$), action-context offset ($\widehat{\pmb{\Psi}}\in \mathbb{R}^{N \times T}$). These outputs are leveraged for the TAL task. We first introduce the TAL baseline using only outputs of \fb. Secondly, we present the contribution of \ac~to the TAL task.

\subsection{Localization Baseline}

The localization baseline uses only outputs of \fb. The temporal action proposals are generated by thresholding ${\pmb{\varphi}}$ with $0.5$. The evaluation (scoring) of temporal action proposals is based on ${\pmb{\Psi}}$.

After obtaining a proposal $\mathbf{P}=[t_s, t_e]$, where $t_s$ and $t_e$ denote the starting and ending snippet indices, respectively. $\mathbf{P}$ is scored by leveraging the Outer-Inner-Contrastive loss~\cite{shou2018autoloc} as
\begin{linenomath}\begin{align}\label{eq:oic}
&s(\mathbf{P}, \mathbf{v})=\textrm{mean}(\mathbf{v}(t_s:t_e))- \nonumber \\
&~~~~~~~~~~~~~~~~~~\textrm{mean}(\langle\mathbf{v}(t_s-\tau:t_s), \mathbf{v}(t_e:t_e+\tau)\rangle),
\end{align}\end{linenomath}
where $\mathbf{v} \in \mathbb{R}^{1 \times T}$ is the sequence for scoring. $\tau=(t_e-t_s)/4$ denotes the inflation length and $\textrm{mean}(\cdot)$ is the averaging function.
Specifically, when locating the $n$-th action category based on ${\pmb{\Psi}}$, we make $\mathbf{v}=\mathbf{v_1}={\pmb{\Psi}}(n,:)$, which is the predictions of the $n$-th action category of all snippets.
After obtaining proposals and their scores, the TAL results are collected.

\subsection{Improving Localization by \ac}\label{sec:loc_ac}

The two critical steps of performing TAL are the generation and evaluation of proposals. The outputs of \ac~can improve both of them. For proposal generation, in addition to thresholding ${\pmb{\varphi}}$ ($P_1$ in Table~\ref{table:ablation_components}), we also perform thresholding step on $\widehat{{\pmb{\varphi}}}_a$ and $\widehat{\pmb{\Psi}}$ ($P_2$ and $P_3$ in Table~\ref{table:ablation_components}). Since $\widehat{\pmb{\varphi}}_a$ and $\widehat{\pmb{\Psi}}$ are both action-aware and less susceptible to the influence of context, the proposals obtained by thresholding them can provide more accurate action boundaries and less context noise.

For proposal evaluation, we can improve the quality of ${\pmb{\Psi}}(n,:)$ to make the scores calculated by Eq.(\ref{eq:oic}) more reliable using $\widehat{\pmb{\Psi}}$. Specifically, we improve ${\pmb{\Psi}}(n,:)$ by suppressing the context and promoting the action as
\begin{equation}\label{eq:fix}
\mathbf{v}_2 = {\pmb{\Psi}}(n,:)+\widehat{\pmb{\Psi}}(n,:).
\end{equation}
By replacing $\mathbf{v}$ with $\mathbf{v}_2$ in Eq.(\ref{eq:oic}), we can evaluate proposals more accurately by alleviating the influence of context.

In summery, the contribution of \ac~to the TAL is reflected in three aspects, \ie, using its outputs ($\widehat{\pmb{\varphi}}_{{a}}$ and $\widehat{\pmb{\Psi}}$) to improve proposal generation ($P_2$ and $P_3$), and using $\widehat{\pmb{\Psi}}$ to improve proposal scoring ($\mathbf{v}_2$). These three aspects are validated in Table~\ref{table:ablation_components}.

\begin{table}[t]
\caption{TAL performance comparison on THUMOS14 test set, in terms of average mAP at IoU thresholds $[0.3 : 0.1 : 0.7]$. Recent works in both fully-supervised and weakly-supervised settings are reported. UNT and I3D represent UntrimmedNet and I3D feature backbones, respectively. ACSNet achieves state-of-the-art performance on both backbones.
Compared to fully-supervised methods, our ACSNet can achieve close or even better performance.
}
\label{table:res_th}
\begin{center}
\resizebox{.48\textwidth}{!}{
\begin{tabular}{c|c|c|p{0.5cm}<{\centering}p{0.5cm}<{\centering}p{0.5cm}<{\centering}p{0.5cm}<{\centering}p{0.5cm}<{\centering}|c}
\hline
\multirow{2}{*}{} & \multirow{2}{*}{Method} & \multirow{2}{*}{\footnotesize{Feature}} & \multicolumn{5}{c|}{mAP@IoU} &  \multirow{2}{*}{AVG}\\
 & & & 0.3 & 0.4 & 0.5 & 0.6 & 0.7 &\\
\hline\hline
\multirow{4}{*}{\rotatebox{90}{\shortstack{Full}}}
&SSN~(\citeyear{zhao2017temporal})  & UNT & 51.9 & 41.0 & 29.8 & 19.6 & 10.7 & 30.6\\
&BSN~(\citeyear{lin2018bsn})   & - & 53.5 & 45.0 & 36.9 & 28.4 & 20.0 & 36.8\\
&MGG~(\citeyear{MGG})   & I3D & 53.9 & 46.8 & 37.4 & 29.5 & 21.3 & 37.8\\
&G-TAD~\citeyear{GTAD}  & - & \textbf{54.5} &\textbf{47.6}	&\textbf{40.2}	&\textbf{30.8}	&\textbf{23.4}	&\textbf{39.3}\\

\hline \hline
\multirow{6}{*}{\rotatebox{90}{\shortstack{Weak}}}
&STPN~(\citeyear{nguyen2017weakly})   & UNT &31.1	&23.5	&16.2	&9.8	&5.1	&17.1\\
&W-TALC~(\citeyear{WTALC})            & UNT &32	&26.0	&18.8	&10.9	&6.2	&18.8\\
&AutoLoc~(\citeyear{shou2018autoloc}) & UNT & 35.8 & 29.0 & 21.2 & 13.4 & 5.8     & 21.0\\
&CleanNet~(\citeyear{lzyiccv})        & UNT & 37.0 & 30.9 & 23.9 & 13.9 & 7.1     & 22.6\\
&\textbf{ACSNet (Ours)}         & UNT & \textbf{40.3} & \textbf{33.8} & \textbf{26.7} & \textbf{16.8} & \textbf{9.2} & \textbf{25.4}\\
\hline \hline
\multirow{9}{*}{\rotatebox{90}{\shortstack{Weak}}}
&STPN~(\citeyear{nguyen2017weakly}) & I3D & 35.5 & 25.8 & 16.9 & 9.9 & 4.3  & 18.5\\
&MAAN~(\citeyear{MAAN_2019_ICLR})   & I3D & 41.1 & 30.6 & 20.3 & 12.0 & 6.9 & 22.2\\
&W-TALC~(\citeyear{WTALC})          & I3D & 40.1 & 31.1 & 22.8 & 14.5 & 7.6 & 23.2\\
&Liu(\citeyear{Liu_2019_CVPR})& I3D & 41.2 & 32.1 & 23.1 & 15.0 & 7.0 & 23.7\\
&BM~(\citeyear{nguyen2019weakly}) & I3D &46.6 &37.5 &26.8 &17.6 &9.0  &27.5\\
&ASSG~(\citeyear{zhang2019adversarial})     & I3D &50.4 &38.7 &25.4 &15.0 &6.6  &27.2\\
&BaSNet~(\citeyear{lee2019background})       & I3D &44.6 &36.0 &27.0 &18.6 &10.4 &27.3\\
&DGAM ~(\citeyear{DGAM})                            & I3D & {46.8} & {38.2} & {28.8} & {19.8} & {11.4} &29.0\\
&\textbf{ACSNet (Ours)}               & I3D &\textbf{51.4}	&\textbf{42.7}	&\textbf{32.4}	&\textbf{22.0}	&\textbf{11.7} &\textbf{32.0}\\
\hline
\end{tabular}
}
\end{center}
\end{table}

\section{Experiments}
In this section, we evaluate the proposed \name~via detailed ablation studies to explore the contribution brought by \ac. Meanwhile, we compare our method with state-of-the-art \wtal~methods and recent fully-supervised TAL methods on two standard benchmarks.
\subsection{Experimental Setting}\label{sec:exp1}

\noindent\textbf{Evaluation Datasets.}
THUMOS14 dataset~\cite{THUMOS14} provides temporal annotations for $20$ action categories, including 200 untrimmed videos from validation set and 213 untrimmed videos from test set.
On average, each video contains $15.4$ action instances and $71.4\%$ frames are non-action background.
Following conventions, the validation and test sets are leveraged for training and testing, respectively.
ActivityNet v1.2 \& v1.3~\cite{caba2015activitynet} provide temporal annotations for $100$ / $200$ action categories, including a training set with $4,819$ / $10,024$ untrimmed videos and a validation set with $2,383$ / $4,926$ untrimmed videos\footnote{\small{In our experiments, there are $4,471$ / $9,937$ and $2,211$ / $4,575$ videos accessible from YouTube in the training and validation set for ActivityNet v1.2 / v1.3, respectively.}}.

\noindent\textbf{Evaluation metric.}
Following the standard evaluation protocol, we evaluate the TAL performance using mean average precision (mAP) values at different levels of IoU thresholds.
Specifically, the IoU threshold sets are $[0.3 : 0.1 : 0.7]$ and $[0.5 : 0.05 : 0.95]$ for THUMOS14 and ActivityNet, respectively.
Both THUMOS14
and ActivityNet
benchmarks provide standard evaluation implementations, which are directly exploited in our experiments for fair comparison.

\begin{table}[t]
	\centering
		\caption{TAL performance comparison on ActivityNet v1.2 and v1.3 validation set, in terms of average mAP at IoU thresholds $[0.5 : 0.05 : 0.95].$ Our result is also comparable to fully-supervised models.}
		\label{table:res_anet}
\resizebox{.48\textwidth}{!}{
\begin{tabular}{c|c|c|ccc|c}
\hline
\multirow{2}{*}{} & \multirow{2}{*}{Method} & \multirow{2}{*}{\footnotesize{1.2 /1.3}} & \multicolumn{3}{c|}{mAP(\%)@IoU} & \multirow{2}{*}{Avg} \\
&  & &  0.5    & 0.75& 0.95  &               \\\hline
\multirow{2}{*}{\rotatebox{90}{\shortstack{Full}}}
& SSN~(\citeyear{zhao2017temporal})          &v1.2& 41.3& 27.0& 6.1  & 26.6          \\
& SSN~(\citeyear{zhao2017temporal})          &v1.3& 39.1& 23.5& 5.5  & 24.0          \\ \hline \hline

\multirow{8}{*}{\rotatebox{90}{\shortstack{Weak}}}
&{AutoLoc}~(\citeyear{shou2018autoloc})            &v1.2 & {27.3} & {15.1} & {3.3} & {16.0} \\
& TSM ~(\citeyear{yu2019temporal})                 &v1.2 & {28.3} & {17.0} & {3.5} & {17.1} \\
& W-TALC ~(\citeyear{WTALC})                      &v1.2 & {37.0} & {12.7} & {1.5} & {18.0} \\
&{\lzynet}  ~(\citeyear{lzyiccv})                  &v1.2 & {37.1} & {20.3} & {5.0} & {21.6} \\
& Liu \etal (\citeyear{Liu_2019_CVPR})            &v1.2 & {36.8} &  22.0  & {5.6} & {22.4} \\
& BaSNet  ~(\citeyear{lee2019background})          &v1.2 & 38.5   & 24.2    & 5.6  & 24.3   \\
& DGAM   ~(\citeyear{DGAM})                        &v1.2 & \textbf{41.0} & {23.5} & 5.3 & {24.4}\\
&\textbf{ACSNet (Ours)}                      &v1.2& {40.1} & \textbf{26.1}& \textbf{6.8} & \textbf{26.0} \\\hline \hline
\multirow{6}{*}{\rotatebox{90}{\shortstack{Weak}}}
& STPN ~(\citeyear{nguyen2017weakly})             &v1.3 & 29.3 & 16.9& 2.6 & -          \\
& TSM    ~(\citeyear{yu2019temporal})               &v1.3 & 30.3 & 19.0& 4.5 & -          \\
& Liu \etal   (\citeyear{Liu_2019_CVPR})           &v1.3 & {34.0} & 20.9 & {5.7} & {21.2} \\
& BM  ~(\citeyear{nguyen2019weakly})       &v1.3 & \textbf{36.4} & 19.2& 2.9 & -          \\
& BaSNet ~(\citeyear{lee2019background})        &v1.3 & 34.5& 22.5& 4.9 & 22.2          \\
& \textbf{ACSNet (Ours)}                      &v1.3 & 36.3 & \textbf{24.2} & \textbf{5.8} & \textbf{23.9}          \\
\hline
\end{tabular}}
\end{table}
\subsection{Comparisons with State-of-the-Art Methods}
As presented in Table~\ref{table:res_th}, the proposed \name~outperforms existing WS-TAL methods in terms of mAPs with all IoU threshold settings on THUMOS14 testing set with significant improvement. Also, the proposed \name~achieves state-of-the-art on ActivityNet v1.2 and v1.3, as presented in Table~\ref{table:res_anet}. However, such performance improvement is not as significant as that on THUMOS14, possibly due to ActivityNet v1.2/v1.3 only has $34.6\%$/$35.7\%$
non-action frames per video on average, while THUMOS14 contains $71.4\%$ on average. With lower non-action ratio, the improvement brought by context suppression could be less significant.

\subsection{Ablation Study}\label{sec:abl}\label{sec:abl_L}

\noindent\textbf{Is Context Really Useful for Classification?}
We assume that the action-context confusion is caused by both action and context can support the classification, due to the high co-occurrence of them.
To validate whether the context snippets estimated by \ac~meet our assumption or not, we collect the foreground/background and action/context snippets as follows. The $t$-th snippet belongs to foreground if $\pmb{\varphi}(t)>0.5$ and otherwise it belongs to background. Among foreground snippets, if $\widehat{\pmb{\varphi}}_{{a}}(t)>0.5$, the $t$-th snippet is assigned as action and otherwise as context. For reference, we also collect all ground truth snippets. Therefore, five snippet sets are collected, noted as $\mathbb{S}_\textrm{fg}$, $\mathbb{S}_\textrm{bg}$, $\mathbb{S}_a$, $\mathbb{S}_c$, and $\mathbb{S}_\textrm{gt}$, respectively.

Regarding the conjuncted snippets as temporal proposals among each set, these snippet sets can be evaluated in both localization and classification tasks, as summarized in Table~\ref{table:context_cls}. For localization, we use the metrics introduced in Section~\ref{sec:exp1} with $\mathbf{v}=\mathbf{v_1}={\pmb{\Psi}}(n,:)$ for proposal evaluation, since ${\pmb{\Psi}}(n,:)$ does not bias on either action or context. For classification, two metrics are adopted, \ie, the average $top1$ classification accuracy ($A_{1}$) and proportion of groundtruth actions defined as
\begin{equation}\label{eq:rate}
R_{z}=\frac{\sum_{t \in \mathbb{S}_{z}} \pmb{\Psi}(n_\textrm{gt},t)}{\sum_{n=1}^{N}\sum_{t \in \mathbb{S}_{z}} \pmb{\Psi}(n,t)}, ~z \in \{ \textrm{fg}, \textrm{bg}, a, c, \textrm{gt}\},
\end{equation}
where $n_\textrm{gt}$ means the groundtruth category and $\pmb{\Psi}(n,t)$ is the $t$-th snippet's classification prediction on the $n$-th class.

As presented in Table~\ref{table:context_cls}, context snippets $\mathbb{S}_c$ contain more useful information compared with $\mathbb{S}_\textrm{bg}$, indicated by the much better classification accuracy. However, in terms of localization task, both $\mathbb{S}_c$ and $\mathbb{S}_c$ perform poorly, which matches our assumption of context, \ie, snippets that can support classification but contain no actual actions.

\begin{table}[t]\small
\caption{
Classification and localization evaluation on different snippet sets on THUMOS14 test set.
\textbf{Classification metric}: Average $top1$ classification accuracy ($A_1$), and proportion of groundtruth actions ($R_z$) defined in Eq.(\ref{eq:rate}). \textbf{Localization metric}: Average mAP under the IoU thresholds from $0.3$ to $0.7$.
}
\begin{center}
\resizebox{.48\textwidth}{!}{
\begin{tabular}{c|cc|ccccc|c}
\hline
\multirow{2}{*}{}&{$A_{1}$}&{$R_{z}$}&\multicolumn{5}{c|}{mAP(\%)@IoU} & \multirow{2}{*}{AVG}\\
&(\%)&(\%)& 0.3 & 0.4 & 0.5 & 0.6 & 0.7 &\\ \hline
$\mathbb{S}_\textrm{gt}$ &91.4&62.4&100&100&100&100&100&100\\ \hline
$\mathbb{S}_\textrm{fg}$ &88.6&59.1&38.3&30.4&21.5&14.4&7.4&22.4\\
$\mathbb{S}_a$           &91.0&61.5&42.4	&34.6	&25.0	&16.7	&9.4	&25.6\\
$\mathbb{S}_c$           &81.0&53.4&0.7&0.3&0.2&0&0&0.2\\
$\mathbb{S}_\textrm{bg}$ &26.7&15.1&0.1&0&0&0&0&0\\ \hline
\end{tabular}
}
\end{center}
\label{table:context_cls}
\end{table}

\begin{table}[t]\small
\caption{
Ablation studies of \name~on THUMOS14 test.
As defined in Section~\ref{sec:abl_L}, the usage of $P_2$/$P_3$/$S_2$ reflect the contribution of ${\widehat{\pmb{\varphi}}_a}$/$\widehat{\pmb{\Psi}}$/$\widehat{\pmb{\Psi}}$ in aspects of proposal generation/generation/evaluation.
$P_2$/$P_3$/$S_2$ take up $33.3\%$/$22.2\%$/$44.5\%$ of the mAP gain upon \#0 ($\alpha:0.4$).
}
\begin{center}
\resizebox{.48\textwidth}{!}{

\begin{tabular}{c|p{0.1cm}<{\centering}p{0.1cm}<{\centering}p{0.2cm}<{\centering}|c|ccccc|c}
\hline
\multirow{2}{*}{Variants}&\multirow{2}{*}{$\!P_1\!$}&\multirow{2}{*}{$\!P_2\!$}&\multirow{2}{*}{$\!P_3\!$} &\multirow{2}{*}{$S_?$} &\multicolumn{5}{c|}{mAP(\%)@IoU} & \multirow{2}{*}{AVG}\\
&&&& & 0.3 & 0.4 & 0.5 & 0.6 & 0.7 &\\ \hline \hline
\#0($\alpha\!:\!0.5$)    &\checkmark&&   &$S_1$  &31.4&23.4&15.8& 9.4&4.8&17.0\\
\#0($\alpha\!:\!0.4$)    &\checkmark&&   &$S_1$  &38.3&30.4&21.5&14.4&7.4&$22.4_{|0.0\%}$\\
\hline \hline

\#1  &&\checkmark&                          &$S_1$ &42.4	&34.6	&25.0	&16.7	&9.4	&$25.6_{|33.3\%}$\\
\#2  &&\checkmark&                          &$S_2$ &49.5	&40.7	&29.3	&19.4	&10.2	&$29.8_{|77.8\%}$\\

\#3  &&&\checkmark                          &$S_2$ &51.6	&42.2	&31.6	&20.6	&10.8	&$31.3_{|92.7\%}$\\
\#4  &&\checkmark&\checkmark                &$S_2$ &\textbf{51.4}	&\textbf{42.7}	&\textbf{32.4}	&\textbf{22.0}	&\textbf{11.7}	&$\textbf{32.0}_{|100\%}$\\
\#5  &\checkmark&\checkmark&\checkmark      &$S_2$ &46.0	&38.5	&{28.4}	&{19.1}	&{9.8}	&28.3\\ \hline \hline
\end{tabular}
}
\end{center}
\label{table:ablation_components}
\end{table}

\noindent\textbf{TAL Contribution of \ac.}
The contribution of the proposed \ac~towards the TAL task is reflected in three aspects as summarized in Section~\ref{sec:loc_ac}.
To validate these three aspects, five ablated variants are evaluated in this section. For the convenience of the discussion, we define the following notations for experiment settings.
For proposal generation settings, $P_1$/$P_2$/$P_3$ are defined as: Thresholding ${\pmb{\varphi}}$/$\widehat{\pmb{\varphi}}_a$/${\widehat{\pmb{\Psi}}}(n,:)$ with $0.5$/$0.5$/$0$ to generate temporal action proposals for all/all/$n$-th action class.
For proposal scoring settings, $S_1$/$S_2$ are defined as: Using $\mathbf{v}_1$/$\mathbf{v}_2$ as the $\mathbf{v}$ in Eq.(\ref{eq:oic}) for proposal evaluation. Therefore, the usage of $P_2$ reflects the contribution of ${\widehat{\pmb{\varphi}}_a}$ in aspects of proposal generation. The usage of $P_3$ and $S_2$ reflect the contribution of $\widehat{\pmb{\Psi}}$ in aspects of proposal generation and evaluation, respectively. The contribution of $P_2$/$P_3$/$S_2$ to TAL is evaluated individually below, as presented in Table~\ref{table:ablation_components}.

With $P_1$ and $S_1$, the \#0 variants are the baseline methods, which depend on \fb~and are non-related to the \ac~. Noted that baselines show super sensitivity towards hyper-parameter $\alpha$, we choose the best one ($\alpha=0.4$) for comparison below. In contrast, all the other ablated variants are with simple average two-stream fusion ($\alpha=0.5$).
Comparison between baseline (\#0) and \#1 shows the contribution solely from $P_2$. Similarly, the contributions solely from $P_3$ and $S_2$ can be validated by the comparisons between \#2 and \#4, \#1 and \#2, respectively.
Quantitatively, $P_2$/$P_3$/$S_2$ take up $33.3\%$/$22.2\%$/$44.5\%$ of the performance gain upon baseline.

Besides, compared with \#4 and \#5, an obvious performance drop is observed, indicating the localization result from \fb~has been burden for the final localization. Without the proposals from \fb, and with the help of ${\widehat{\pmb{\varphi}}_a}$ and $\widehat{\pmb{\Psi}}$ on proposal generation and evaluation, ``\#4'' achieves the best localization performance.

\section{Conclusions}
We propose an \name~for weakly-supervised temporal action localization, which can separate action and context with only video-level categorical labels. This is achieved by characterizing foreground/action/context as combinations of positive and negative latent compositions. \name~significantly outperforms existing \wtal~methods on three standard datasets, \ie, THUMOS14, ActivityNet v1.2 and v1.3. Moreover, \name~achieves competitive performance even compared with recent fully-supervised TAL methods. Experimental results validate the significance of action-context separation and the superiority of the proposed pipeline.

\section{Acknowledgments}
This work was supported partly by National Key R\&D Program of China Grant 2018AAA0101400, NSFC Grants 61629301, 61773312, and 61976171, China Postdoctoral Science Foundation Grant 2019M653642, Young Elite Scientists Sponsorship Program by CAST Grant 2018QNRC001, and Natural Science Foundation of Shaanxi Grant 2020JQ-069.

\bibliography{AAAI_preprint}
\end{document}